\documentclass[twoside,11pt]{article}

\usepackage{jmlr2e}

\usepackage{graphicx}  
\usepackage[round]{natbib}

\usepackage{url}  

\usepackage[dvipsnames]{xcolor}

\usepackage[backref=page]{hyperref}
\hypersetup{
  citecolor = NavyBlue,
  colorlinks = true
}
\usepackage{backref}

\usepackage{subcaption}
\usepackage[switch,columnwise]{lineno}
\usepackage[lined, boxed, ruled, commentsnumbered, noend]{algorithm2e}

\usepackage{amssymb,amsmath}
\usepackage{mathtools}
\usepackage{xspace}

\usepackage[makeroom]{cancel}

\SetCommentSty{mycommfont}

\newcommand{\denselist}{
  \itemsep 0pt \topsep-8pt\partopsep-8pt
}

\DeclareMathOperator*{\argmax}{arg\,max}

\newcommand\floor[1]{\left\lfloor #1 \right \rfloor}

\newcommand{\defref}[1]{Definition~\ref{#1}}

\newcommand{\figref}[1]{Fig.~\ref{#1}}

\newcommand{\secref}[1]{\S\ref{#1}}
\newcommand{\thmref}[1]{Theorem~\ref{#1}}
\newcommand{\corref}[1]{Corollary~\ref{#1}}

\newcommand{\lnref}[1]{Line~\ref{#1}}
\newcommand{\algref}[1]{Algorithm~\ref{#1}}

\newcommand{\algname}{\textsf{MF-MI-Greedy}\xspace}
\newcommand{\sfgpopt}{\textsf{SF-GP-OPT}\xspace}
\newcommand{\explorelf}{\textsf{Explore-LF}\xspace}

\newcommand{\sfgpucb}{\textsf{GP-UCB}\xspace}
\newcommand{\mfgpucb}{\textsf{MF-GP-UCB}\xspace}
\newcommand{\sfmves}{\textsf{MVES}\xspace}
\newcommand{\sfest}{\textsf{EST}\xspace}

\newcommand{\sfgpmi}{\textsf{GP-MI}\xspace}

\newcommand{\nan}{\textsf{null}\xspace}

\newcommand{\mean}{\mu}
\newcommand{\std}{\sigma}
\newcommand{\cov}{k}
\newcommand{\bcov}{\bk}
\newcommand{\Cov}{K}

\usepackage{dsfont}

\newcommand{\expct}[1]{\mathbb{E}\left[#1\right]}
\newcommand{\expctover}[2]{\mathbb{E}_{#1}\!\left[#2\right]}

\newcommand{\cE}{{\mathcal{E}}}
\newcommand{\cS}{{\mathcal{S}}}

\newcommand{\cL}{{\mathcal{L}}}

\newcommand{\bk}{{\mathbf{k}}}

\newcommand{\by}{{\mathbf{y}}}

\newcommand{\paren} [1] {\ensuremath{ \left( {#1} \right) }}

\newcommand{\littleO}[1]{\ensuremath{o\paren{#1}}}
\newcommand{\bigO}[1]{\ensuremath{O\paren{#1}}}

\newcommand{\bigOmega}[1]{\ensuremath{\Omega\paren{#1}}}

\newcommand{\integers}{\ensuremath{\mathbb{Z}}}

\newcommand{\reals}{\ensuremath{\mathbb{R}}}

\newcommand{\GP}[1]{\text{GP}\paren{#1}} 

\newcommand{\reward}{q} 

\newcommand{\fidelity}{f} 
\newcommand{\utility}{f} 

\newcommand{\exDom}{\mathcal{X}} 

\newcommand{\ep}{\cE} 
\newcommand{\ex}{x} 
\newcommand{\obs}{y} 
\newcommand{\bobs}{\by} 

\newcommand{\selected}{\cS} 
\newcommand{\epselected}{\cE} 
\newcommand{\eplow}{\cL} 

\newcommand{\epselectedat}[1]{\cE^{\paren{#1}}} 
\newcommand{\eplowat}[1]{\cL^{\paren{#1}}} 
\newcommand{\tarselectedat}[1]{\action{\ex, \targetfid}^{\paren{#1}}} 

\newcommand{\noise}{\varepsilon}

\newcommand{\action}[1]{\langle #1\rangle} 
\newcommand{\step}{t}
\newcommand{\stepiter}{\tau}

\newcommand{\fid}{\ell}
\newcommand{\targetfid}{m} 

\newcommand{\normal}{\mathcal{N}}

\newcommand{\tarf}{f_\targetfid} 
\newcommand{\regret}{r} 
\newcommand{\cumreg}{R} 

\newcommand{\budget}{\Lambda} 
\newcommand{\Cost}{\Lambda} 

\newcommand{\costof}[1]{\lambda_{#1}} 
\newcommand{\epcost}{\Lambda_{\ep}} 

\newcommand{\epcostat}[1]{\Lambda_\ep^{\paren{#1}}}
\newcommand{\eplowcostat}[1]{\Lambda_\eplow^{\paren{#1}}}

\DeclarePairedDelimiterX\parencond[2]{(}{)}{#1 \;\delimsize\vert\; #2}
\DeclarePairedDelimiterX\bracketcond[2]{\{}{\}}{#1 \;\delimsize\vert\; #2}

\newcommand{\condinfgain}[2]{\mathbb{I}\parencond*{#1; \tarf}{#2}} 
\newcommand{\condentropy}[2]{\mathbb{H}\parencond*{#1}{#2}} 

\newcommand{\constgeneral}{C}

\newcommand{\policy}{\pi} 

\begin{document}
\title{A General Framework for Multi-fidelity Bayesian Optimization with Gaussian Processes}

\author{\name Jialin Song \email jssong@caltech.edu \\
  \addr California Institute of Technology \\ Pasadena, CA, USA
  \AND
  \name Yuxin Chen \email chenyux@caltech.edu \\
  \addr California Institute of Technology \\ Pasadena, CA, USA
  \AND
  \name Yisong Yue  \email yyue@caltech.edu\\
  \addr California Institute of Technology \\ Pasadena, CA, USA
}


\maketitle






\begin{abstract}







  How can we efficiently gather information to optimize an unknown function, when presented with multiple, mutually dependent information sources with different costs? For example, when optimizing a robotic system, intelligently trading off computer simulations and real robot testings can lead to significant savings. Existing methods, such as multi-fidelity GP-UCB or Entropy Search-based approaches, either make simplistic assumptions on the interaction among different fidelities or use simple heuristics that lack theoretical guarantees. In this paper, we study multi-fidelity Bayesian optimization with complex structural dependencies among multiple outputs, and propose \algname, a principled algorithmic framework for addressing this problem. In particular, we model different fidelities using additive Gaussian processes based on shared latent structures with the target function. Then we use cost-sensitive mutual information gain for efficient Bayesian global optimization. We propose a simple notion of regret which incorporates the cost of different fidelities, and prove that \algname achieves low regret. We demonstrate the strong empirical performance of our algorithm on both synthetic and real-world datasets.
\end{abstract}


\section{Introduction}

\begin{figure*}[t]
  \centering
  \begin{subfigure}[b]{.49\textwidth}
    \centering
    {
      \includegraphics[trim={0pt 0pt 0pt 0pt}, width=\textwidth]{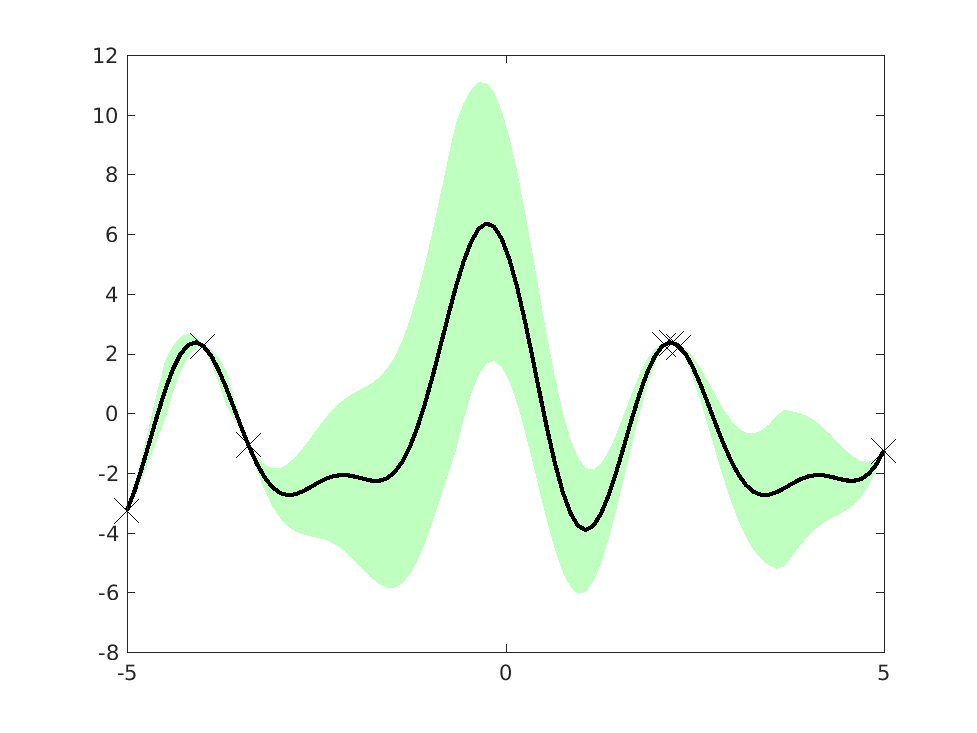}
      \caption{Only querying target fidelity function.}
      \label{fig:intro:mf:step1}
    }
  \end{subfigure}
  \begin{subfigure}[b]{.49\textwidth}
    \centering
    {
      \includegraphics[trim={0pt 0pt 0pt 0pt}, width=\textwidth]{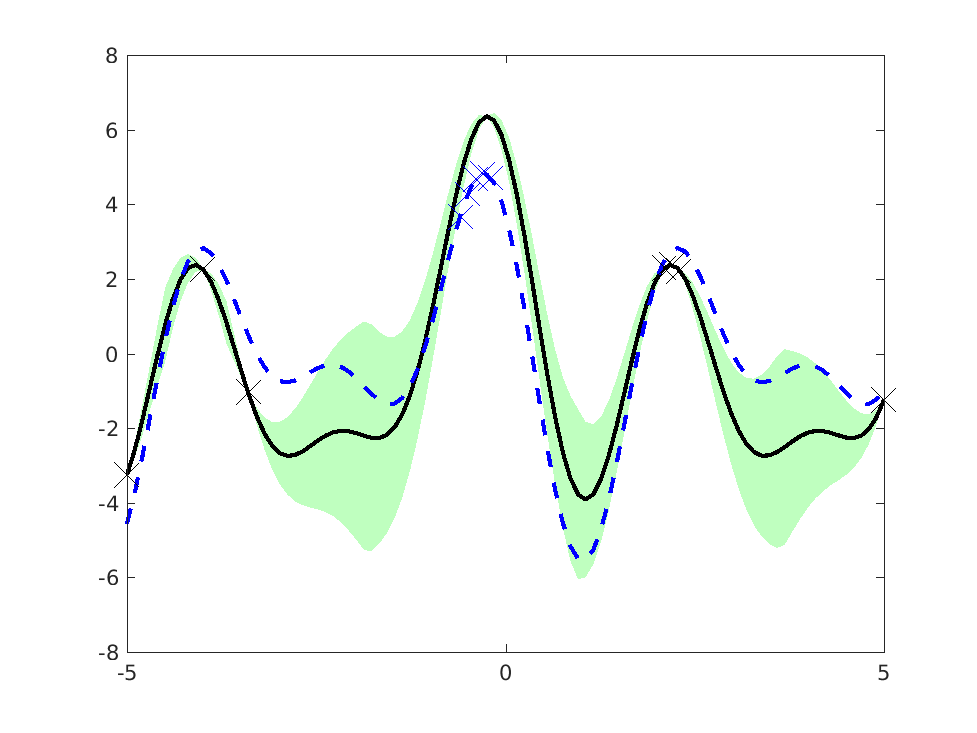}
      \caption{Querying both target and a lower fidelity.}
      \label{fig:intro:mf:step2}
    }
  \end{subfigure}
  \caption{Benefit from multi-fidelity Bayesian optimization. The left panel shows normal single fidelity Bayesian optimization where locations near a query point (crosses) have low uncertainty. When there is a lower fidelity cheaper approximation in the right panel, by querying a large number of points of the lower fidelity function, the uncertainty in the target fidelity can also be reduced significantly.}
  \label{fig:intro:multifidelity}
\end{figure*}

Optimizing an unknown function that is expensive to evaluate is a common problem in real applications. Examples include experimental design for protein engineering, where chemists need to synthesize designed amino acid sequences and then test whether they satisfy certain properties \citep{romero2013navigating}; or black-box optimization for material science, where scientists need to run extensive computational experiments at various levels of accuracy to find the optimal material design structure 
\citep{fleischman2017hyper}.
Conducting real experiments could be labor-intensive and time-consuming. In practice, we would like to look for alternative ways to gather information so that we can make the most effective use of real experiments that we do conduct. A natural candidate is computer simulation \citep{van1990computer}, which tends to be less time consuming but produces less accurate results. For example, computer simulation is ubiquitous in robotic applications, e.g. we test a control policy first in simulation before deploying it in a real physical system \citep{marco2017virtual}.

The central challenge in efficiently using multiple sources of information is captured in the general framework of multi-fidelity optimization \citep{forrester2007multi, kandasamy2016gaussian, kandasamy2017multi, marco2017virtual, sen2018mfpdoo} where multiple functions with varying degrees of accuracy and costs can be effectively leveraged to provide the maximal amount of information. However, strict assumptions, such as requiring strict relations between the quality and the cost of a lower fidelity function, and two-stage query selection criteria (cf. \secref{sec:related:gpbo}) are likely to limit their practical use and lead to sub-optimal selections. 

In this paper, we propose a general and principled multi-fidelity Bayesian optimization framework \algname (Multi-fidelity Mutual Information Greedy) that prioritizes maximizing the amount of mutual information gathered across fidelity levels. Figure \ref{fig:intro:multifidelity} captures the intuition of maximizing mutual information. Gathering information from lower fidelity also conveys information on the target fidelity. We make this idea concrete in \secref{sec:alg}. Our method improves upon prior work on multi-fidelity Bayesian optimization by establishing explicit connections across fidelity levels to enable joint posterior updates and hyperparameter optimization. In summary, our contributions in this paper are

\begin{itemize}\denselist
\item We study multi-fidelity Bayesian optimization with complex structural dependencies among multiple outputs (\secref{sec:statement}), and propose \algname, a principled algorithmic framework for addressing this problem (\secref{sec:alg}).
\item We propose a simple notion of regret which incorporates the cost of different fidelities, and prove that \algname achieves low regret (\secref{sec:analysis}).
\item We demonstrate the empirical performance of \algname on both simulated and real-world datasets (\secref{sec:exp}).
\end{itemize}


\section{Background and Related Work}
In this section, we review related work on Bayesian optimization with 
Gaussian processes.
\subsection{Background on Gaussian Processes}
Gaussian process \citep{rasmussen:williams:2006} models an infinite collection of random variables, each indexed by an $\ex \in \exDom$, such that every finite subset of random variables has a multivariate Gaussian distribution. The GP distribution $\GP{\mean(\ex), \cov(\ex, \ex')}$ is a joint Gaussian distribution over all those (infinitely many) random variables specified by its mean $\mean(\ex) = \expct{\utility(\ex)}$ and covariance (also known as \emph{kernel}) function $$\cov(\ex, \ex') = \expct{\paren{\utility(\ex) - \mean(\ex)}\paren{\utility(\ex') - \mean(\ex')}}.$$

A key advantage of GP is that it is very efficient to perform inference.
Assume that $\utility \sim \GP{\mean(\ex), \cov(\ex, \ex')}$ is a sample from the GP distribution, and that $\obs = \utility(\ex) + \noise(\ex)$ is a noisy observation of the function value $\utility(\ex)$. Here, the noise $\noise(\ex)\sim \normal(0, \sigma^2(\ex))$ could depend on the input $\ex$. Suppose that we have selected $\selected \subseteq \exDom$ and received $\bobs_\selected = [\utility(\ex_i) + \noise(\ex_i)]_{\ex_i\in\selected}$. We can obtain the posterior mean $\mean_\selected(\ex)$ and covariance $\cov_\selected{(\ex, \ex')}$ of the function through the covariance matrix $\Cov_\selected = [\cov(\ex_i, \ex_j) + \delta_{ij}\sigma^2(\ex_i)]_{\ex_i, \ex_j \in \selected}$ and $\bcov_\selected(\ex) = [\cov(\ex_i, \ex)]_{\ex_i \in \selected}$:
\begin{align}
  \label{eq:gp_posterior}
  \mean_\selected(\ex) &= \mean(\ex) + \bcov_\selected(\ex)^\intercal\Cov_\selected^{-1} \bobs_\selected\\
  \cov_\selected{(\ex, \ex')} &= \cov(\ex, \ex') - \bcov_\selected(\ex)^\intercal\Cov_\selected^{-1} \bcov_\selected(\ex')
\end{align}
where $\delta_{ij}$ is the Kronecker delta function.

\subsection{Bayesian Optimization via Gaussian Processes}\label{sec:related:gpbo}
\paragraph{Single-fidelity Gaussian Process optimization}
Optimizing an unknown and noisy function is a common task in Bayesian optimization. In real applications, such functions tend to be expensive to evaluate, for example tuning hyperparameters for deep learning models \citep{snoek2012practical}, so the number of queries should be minimized.  As a way to model the unknown function, Gaussian process (GP) \citep{rasmussen:williams:2006} is an expressive and flexible tool to model a large class of functions. A classical method for Bayesian optimization with GPs is \sfgpucb \citep{srinivas10gaussian} which treats Bayesian optimization as a multi-armed bandit problem and proposes an upper-confidence bound based algorithm for query selections. The authors provide a theoretical bound on the cumulative regret that is connected with the amount of mutual information gained through the queries.
\citep{contal2014gaussian} directly incorporates mutual information into the UCB framework and demonstrated the empirical value of their method.

Entropy search \citep{hennig2012entropy} represents another class of GP-based Bayesian optimization approach. Its main idea is to directly search for the global optimum of an unknown function through queries. Each query point is selected based on its informativeness in learning the location for the function optimum. Predictive entropy search \citep{hernandez2014predictive} addresses some computational issues from entropy search by maximizing the expected information gain with respect to the location of the global optimum. Max-value entropy search \citep{wang2016optimization,wang2017max} approaches the task of searching the global optimum differently. Instead of searching for the location of the global optimum, it looks for the value of the global optimum. This effectively avoids issues related to the dimension of the search space and the authors are able to provide regret bound analysis that the previous two entropy search methods lack.

A computational consideration for learning with GPs concerns with optimizing specific kernels used to model the covariance structures of GPs. As this optimization task depends on the dimension of feature space, approximation methods are needed to speed up the learning process. Random Fourier features \citep{rahimi2008random} are efficient tools for dimension reduction and are employed in GP regression tasks \citep{lazaro2010sparse}. As elaborated on in \secref{sec:alg}, our algorithmic framework offers the flexibility of choosing among different single-fidelity optimization approaches as a subroutine, so that one can take advantage of these computational and approximation algorithms for efficient optimization. 

\paragraph{Multi-output Gaussian Process}
Sometimes it is desirable to model multiple correlated outputs with Gaussian processes. Most GP-based multi-output models create correlated outputs by mixing a set of independent latent processes. A simple form of such a mixing scheme is the linear model of coregionalization \citep{teh2005semiparametric,bonilla2008multi} where each output is modeled as a linear combination of latent GPs with fixed coefficients. The dependencies among outputs are captured by sharing those latent GPs. More complex structures can be captured by a linear combination of GPs with input-dependent coefficients \citep{wilson2012gaussian}, shared inducing variables \citep{nguyen2014collaborative}, or convolved process \citep{boyle2005dependent,alvarez2009sparse}. In comparison with single fidelity/output GPs, multi-output GP often requires more sophisticated approximate models for efficient optimization (e.g., using inducing points \citep{snelson2007local} to reduce the storage and computational complexity, and variational inference approaches to approximate the posterior of the latent processes \citep{titsias2009variational,nguyen2014collaborative}). While the analysis of our framework is not limited to a fixed structural assumption in modeling the joint distribution among multiple outputs, for efficiency concern, we use a simple, additive model between multiple fidelity outputs in our experiments (cf. \secref{sec:exp:setup}) to demonstrate the effectiveness of the optimization framework.

\paragraph{Multi-fidelity Bayesian optimization}
Multi-fidelity optimization is a general framework that captures the trade-off between cheap low quality and expensive high quality data. Recently, there have been several works on using GPs to model functions of different fidelity levels. Recursive co-kriging \citep{forrester2007multi,le2014recursive} consider an autoregressive model for multi-fidelity GP regression, which assumes that the higher fidelity consists of a lower fidelity term and an \emph{independent} GP term which models the systematic error for approximating the higher-fidelity output. Therefore, one can model cross-covariance between the high-fidelity and low-fidelity functions using the covariance of the lower fidelity function only. Virtual vs Real \citep{marco2017virtual} extends this idea to Bayesian optimization. The authors consider a two-fidelity setting (i.e., virtual simulation and real system experiments), where they model the correlation between the two fidelities through co-kriging, and then apply entropy search (ES) to optimize the target output.
\citet{zhang2017mfpes} model the dependencies between different fidelities with convolved Gaussian processes \citep{alvarez2009sparse}, and then apply predictive entropy search (PES) \citep{hernandez2014predictive} to efficient exploration. Although both the ES and multi-fidelity PES heuristics have shown promising empirical results on some datasets, little is known about their theoretical performance.

Recently, \citet{kandasamy2016gaussian} proposed Multi-fidelity GP-UCB (\mfgpucb), a principled framework for multi-fidelity Bayesian optimization with Gaussian processes. In a followup work \citep{kandasamy2017multi,sen2018mfpdoo}, the authors address the disconnect issue by considering a continuous fidelity space and performing joint updates to effectively share information among different fidelity levels. 

In contrast to our general assumption on the joint distribution between different fidelities, \citet{kandasamy2016gaussian} and \citet{sen2018mfpdoo}
assume a specific structure over multiple fidelities, where the cost of each lower fidelity is determined according to the maximal approximation error in function value when compared with the target fidelity. \citet{kandasamy2017multi} consider a two-stage optimization process, where the action and the fidelity level are selected in two separate stages. We note that this procedure may lead to non-intuitive choices of queries: For example, in a pessimistic case where the low fidelity only differs from the target fidelity by a constant shift, their algorithm is likely to focus only on querying the target fidelity actions even though the low fidelity is as useful as the target fidelity. In contrast, as described in \secref{sec:alg}, our algorithm jointly selects a query point and a fidelity level so such sub-optimality can be avoided.

\section{Problem Statement}\label{sec:statement}
\looseness -1 We now introduce useful notations and formally state the problem studied in this paper.
\paragraph{Payoff function and auxiliary functions} Consider the problem of maximizing an unknown payoff function $\utility_\targetfid: \exDom \rightarrow \reals$.
We can probe the function $\utility_\targetfid$ by directly querying it at some $\ex \in \exDom$ and obtaining a noisy observation $\obs_{\action{\ex, \targetfid}}=\utility_\targetfid(\ex) + \noise(x)$, where $\noise(x) \sim \normal(0, \std^2)$ denotes i.i.d. Gaussian white noise. 
In addition to the payoff function $\utility_\targetfid$, we are also given access to oracle calls to some unknown auxiliary functions $\utility_1, \dots, \utility_{\targetfid-1}: \exDom \rightarrow \reals$; similarly, we obtain a noisy observation $\obs_{\action{\ex, \fid}}=\utility_\fid(\ex) + \noise$ when querying $\utility_\fid$ at $\ex$. Here, each $\utility_\fid$ could be viewed as a low-fidelity version of $\utility_\targetfid$ for $\ell<\targetfid$. For example, if $\utility_\targetfid(\ex)$ represents the actual reward obtained by running a real physical system with input $\ex$, then $\utility_\ell(\ex)$ may represent the simulated payoff from a numerical simulator at fidelity level $\fid$.

\paragraph{Joint distribution on multiple fidelities}

We assume that multiple fidelities $\{\utility_\fid\}_{\fid\in[m]}$ 
are mutually dependent through some fixed, (possibly) unknown joint probability distribution $\mathbb{P}[\utility_1, \dots, \utility_\targetfid]$. 
In particular,
we model $\mathbb{P}$ with a multiple output Gaussian process; hence the marginal distribution on each fidelity is a separate GP, i.e., $\forall \fid\in[\targetfid],\ \utility_\fid \sim \GP{\mean_\fid(\ex), \cov_\fid(\ex, \ex')}$, where $\mean_\fid, \cov_\fid$ specify the (prior) mean and covariance at fidelity level $\fid$. 

\paragraph{Action, reward, and cost} Let us use $\action{\ex, \fid}$ to denote the action of querying $\utility_\fid$ at $\ex$. Each action $\action{\ex, \fid}$ incurs a cost of $\costof{\fid}$, and achieves a reward
\begin{align}
  \label{eq:reward}
  \reward(\action{\ex, \fid}) =
  \begin{cases}
    \utility_\targetfid(\ex) & \text{if}~\fid = \targetfid\\
    \reward_{\min} & \text{o.w.}
  \end{cases}
\end{align}
That is, performing $\action{\ex, \targetfid}$ 
(at the target fidelity) achieves a reward $\utility_\targetfid(\ex)$. We receive the minimal immediate reward $\reward_{\min}$ with lower fidelity actions $\action{\ex, \fid}$ for $\fid < \targetfid$, even though it may provide some information about $\utility_\targetfid$ and could thus lead to more informed decisions in the future. W.l.o.g., we assume that $\max_{\ex} \tarf(\ex) \geq 0$, and $\reward_{\min} \equiv 0$.



\paragraph{Policy} 
Let us encode an adaptive strategy for picking actions as a policy $\policy$. In words, a policy specifies which action to perform next, based on the actions picked so far and the corresponding observations. We consider policies with a fixed budget $\budget$. Upon termination, $\policy$ returns a sequence of actions $\selected_\policy$, such that $\sum_{\action{\ex, \fid} \in \selected{\policy}} \costof{\fid} \leq \budget$. Note that for a given policy $\policy$, the sequence $\selected_\policy$ is random, dependent on the joint distribution $\mathbb{P}$ and the (random) observations of the selected actions.

\paragraph{Objective} Given a budget $\budget$ on $\policy$, our goal is to maximize the expected cumulative reward, so as to identify an action $\action{\ex, \targetfid}$ with performance close to $\ex^*=\max_{\ex \in \exDom} \utility_\targetfid(\ex)$ as rapidly as possible. Formally, we seek
\begin{align}
  \label{eq:maxreward}
  \policy^* = \argmax_{\policy: \sum_{\action{\ex, \fid} \in \selected{\policy}} \costof{\fid} \leq \budget} \expctover{\selected_\policy}{\sum_{\action{\ex, \fid} \in \selected_\policy} \reward(\action{\ex, \fid})}
\end{align}

\paragraph{Remarks}
Problem~\ref{eq:maxreward} strictly generalizes the optimal value of information (VoI) problem \citep{chen15submodular} to the online setting. To see this, consider the scenario where $\costof{\targetfid} \gg \costof{\fid}$ for $\fid < \targetfid$, and $\budget \in (\costof{\targetfid}, 2\costof{\targetfid})$. To achieve a non-zero reward, a policy must pick $\action{\ex, \targetfid}$ as the last action before exhausting the budget $\budget$. Therefore, our goal becomes to adaptively pick lower fidelity actions that are the most informative about $\ex^*$ under budget $\budget-\costof{\targetfid}$, which reduces to the optimal VoI problem. 


\section{The Multi-fidelity BO Framework} \label{sec:alg}
We now present \algname, a general framework for multi-fidelity Gaussian process optimization. In a nutshell, \algname attempts to balance the ``exploratory'' low-fidelity actions and the more expensive target fidelity actions, based on how much information (per unit cost) these actions could provide about the target fidelity function. 
Concretely, \algname proceeds in rounds under a given budget $\budget$. Each round can be divided into two phases: (i) an exploration (i.e., information gathering) phase, where the algorithm focuses on exploring the low fidelity actions, and (ii) an optimization phase, where the algorithm tries to optimize the payoff function $\tarf$ by performing an action $\action{\ex, \targetfid}$ at the target fidelity. The pseudo-code of the algorithm is provided in \algref{alg:main}.

\begin{algorithm}[t]
  \nl {\bf Input}: {Total budget $\budget$; cost $\costof{i}$ for all fidelities $i \in [\targetfid]$}; joint GP (prior) distribution on $\{\fidelity_i, \noise_i\}_{i\in[\targetfid]}$\\ 
  \Begin{
    \nl $\selected \leftarrow \emptyset$ \\
    \nl $B \leftarrow \budget$ \tcc*{initialize remainig budget}
    \While{$B > 0$} 
    {
      \tcc{explore with low fidelity}
      \nl $\eplow \leftarrow$ \explorelf $\paren{B, [\costof{\fid}], \GP{\{\utility_\fid, \noise_\fid\}_{\fid\in[\targetfid]}}, \selected}$ \\
      \tcc{select target fidelity}
      \nl $\ex^* \leftarrow \sfgpopt(\GP{\{\tarf, \noise_\targetfid\}}, \bobs_{\selected \cup \eplow})$ \label{ln:alg:main:sfgpout}\\
      \nl $\selected \leftarrow \selected \cup \eplow \cup \{\action{\ex^*,\targetfid}\}$\\
      \nl $B \leftarrow \budget - \Cost_\selected$ \tcc*{update remaining budget}
    }
    \nl {\bf Output}: Optimizer of the target function $\tarf$ \\
  }
  \caption{Multi-fidelity Mutual Information Greedy Optimization (\algname)}\label{alg:main}
\end{algorithm}

\paragraph{Exploration phase}
A key challenge in designing the algorithm is to decide when to stop exploration (or equivalently, to invoke the optimization phase). Note that this is analogous to the exploration-exploitation dilemma in the classical single-fidelity Bayesian optimization problems; the difference is that in the multi-fidelity setting, we have a more distinctive notion of ``exploration'', and a more complicated structure of the action space (i.e., each exploration phase corresponds to picking a set of low fidelity actions). Furthermore, note that there is no explicit measurement of the relative ``quality'' of a low fidelity action, as they all have uniform reward by our modeling assumption (c.f. Eq.~\eqref{eq:reward}); hence we need to design a proper heuristic to keep track of the progress of exploration.

We consider an information-theoretic selection criterion for picking low fidelity actions. The quality of a low fidelity action $\action{\ex, \fid}$ is captured by the \emph{information gain}, defined as the amount of entropy reduction in the posterior distribution of the target payoff function\footnote{An alternative, more aggressive information measurement is the information gain over the \emph{optimizer} of the target function $\tarf$ \citep{hennig2012entropy}, i.e., $\mathbb{I}\parencond*{\obs_{\action{\ex,\fid}}; \argmax_{\ex}\tarf(\ex)}{\bobs_\selected}$, or the \emph{optimal value} of $\tarf$ \citep{wang2017max}, i.e., $\mathbb{I}\parencond*{\obs_{\action{\ex,\fid}}; \max_{\ex}\tarf(\ex)}{\bobs_\selected}$.
}: $\condinfgain{\obs_{\action{\ex, \fid}} }{\bobs_\selected} = \condentropy{\obs_{\action{\ex, \fid}} }{\bobs_\selected} - \condentropy{\obs_{\action{\ex, \fid}} }{\tarf, \bobs_\selected}$. Here, $\selected$ denotes the set of previously selected actions, and $\bobs_{\selected}$ denote the observation history. 
Given an exploration budget $B$, our objective for a single exploration phase is to find a set of low-fidelity actions, which are (i) maximally informative about the target function, (ii) better than the best action on the target fidelity when considering the information gain per unit cost (otherwise, one would rather pick the target fidelity action to trade off exploration and exploitation), and (iii) not overly aggressive in terms of exploration (since we would also like to reserve a certain budget for performing target fidelity actions to gain reward).

Finding the optimal set of actions satisfying the above design principles is computationally prohibitive, as it requires us to search through a combinatorial (for finite discrete domains) or even infinite (for continuous domains) space. In \algref{alg:explorelf}, we introduce \explorelf, a key subroutine of \algname, for efficient exploration on low fidelities. At each step, \explorelf takes a greedy step w.r.t. the benefit-cost ratio over all actions. To ensure that the algorithm does not  explore excessively, we consider the following stopping conditions: (i) when the budget is exhausted (\lnref{alg:explorelf:ln:budget}), (ii) when a single target fidelity action is better than all the low fidelity actions in terms of the benefit-cost ratio (\lnref{alg:explorelf:ln:targetcheck}), and (iii) when the cumulative benefit-cost ratio is small (\lnref{alg:explorelf:ln:infgaincheck}). Here, the parameter $\beta$ is set to be  $\bigOmega{\frac{1}{\sqrt{B}}}$ to ensure low regret, and we defer the detailed discussion of the choice of $\beta$ to \secref{sec:analysis:regret}.

\begin{algorithm}[t]
  \nl {\bf Input}: {Exploration budget $B$; cost $[\costof{\fid}]_{\fid \in [\targetfid]}$; joint GP distribution on $\{\fidelity_i, \noise_i\}_{i\in[\targetfid]}$; previously selected items $\selected$}\\ 
  \Begin{
    \nl $\eplow \leftarrow \emptyset$ \tcc*{selected actions}
    \nl $\Cost_\eplow \leftarrow 0$ \tcc*{cost of selected actions}
    \nl $\beta \leftarrow \frac{1}{\alpha(B)}$ \tcc*{threshold (c.f. \thmref{thm:general:datadependent-bound})}
    \While{true}
    {
      \tcc{greedy benefit-cost ratio}
      \nl $\action{\ex^*, \fid^*} \leftarrow
      \argmax_{\action{\ex, \fid}: \costof{\fid} \leq B-\Cost_\eplow - \costof{\targetfid}} \frac{\condinfgain{\obs_{\action{\ex, \fid}}}{\bobs_{\selected \cup \eplow}}}{\costof{\fid}}$ \\
      \If{$\fid^* = \nan$}
      {
        \nl break \label{alg:explorelf:ln:budget}
        \tcc*{budget exhausted}
      }
      \If{$\fid^* = \targetfid$}
      {
        \nl break \label{alg:explorelf:ln:targetcheck}
        \tcc*{worse than target}
      }
      \ElseIf{ $\frac{\condinfgain{\bobs_{\eplow \cup \{ \action{\ex^*, \fid^*} \} } }{\bobs_\selected}} {\paren{\Cost_\eplow + \costof{\fid^*}}} < \beta$}
      {
        \nl \label{alg:explorelf:ln:infgaincheck} break \tcc*{low cumulative ratio}
      }
      \Else
      {
        \nl $\eplow \leftarrow \eplow \cup \{\action{\ex^*,\fid^*}\}$ \\
        \nl $\Cost_\eplow \leftarrow \Cost_\eplow + \costof{\fid^*}$
      }
    }
  }
  \nl {\bf Output}: Selected set of items $\eplow$ from lower fidelities \\
  \caption{\explorelf: Explore low fidelities}\label{alg:explorelf}
\end{algorithm}

\paragraph{Optimization phase} At the end of the exploration phase, \algname updates the posterior distribution of the joint GP using the full observation history, 
and searches for a target fidelity action via the (single-fidelity) GP optimization subroutine \sfgpopt (\lnref{ln:alg:main:sfgpout}). 
Here, \sfgpopt could be \emph{any} off-the-shelf Bayesian optimization algorithm, such as \sfgpucb \citep{srinivas10gaussian}, \sfgpmi \citep{contal2014gaussian}, \sfest \citep{wang2016optimization} and \sfmves \citep{wang2017max}, etc. Different from the previous exploration phase which seeks an informative set of low fidelity actions, the GP optimization subroutine aims to trade off exploration and exploitation on the target fidelity, and outputs a single action at each round. %
\algname then proceeds to the next round until it exhausts the preset budget and eventually outputs an estimator of the target function optimizer.



\section{Theoretical Analysis}\label{sec:analysis}
In this section, we investigate the theoretical behavior of \algname. We first introduce an intuitive notion of regret for the multi-fidelity setting, and then state our main theoretical results.
\subsection{Multi-fidelity Regret}
\begin{definition}[Episode] 
  Let $\step\in \integers$ be any integer. We call a sequence of items $\ep = \{\langle \ex_1, \fid_1 \rangle, \dots, \langle \ex_\step, \fid_\step \rangle\}$ an \emph{episode}, if $\forall \stepiter < \step:\ \fid_\stepiter < \targetfid$ and $\fid_\step = \targetfid$.
\end{definition}
In words, only the last action of an episode is from the target fidelity $\tarf$ and all remaining actions are from lower fidelities. We now define a simple notion of regret for an episode.
\begin{definition}[Episode regret]
  The regret of an episode $\ep = \{\action{\ex_1, \fid_1}, \dots, \action{\ex_{\step}, \targetfid \rangle}\}$ is
  \begin{align}
    \regret(\ep) = \frac{\epcost}{\costof{\targetfid}} \tarf^* - \reward(\ep)
  \end{align}
  where $\epcost := \sum_{\langle \ex,\fid \rangle\in \ep} \costof{\fid}$ is the total cost of episode $\ep$, and $\reward(\ep) := \tarf\left(x_{\step}\right)$ denotes the reward value of the last action on the target fidelity.
\end{definition}

Suppose we run policy $\policy$ under budget $\budget$ and select a sequence of actions $\selected_\policy$. One can represent $\selected_\policy$ using multiple episodes $\selected_\policy = \{\epselectedat{1}, \dots, \epselectedat{k}\}$,
where $\epselectedat{j} = \eplowat{j} \cup \{\tarselectedat{j}\}$ denotes the sequence of low fidelity actions $\eplowat{j}$ and target fidelity action $\tarselectedat{j}$ selected at the $j^{\text{th}}$ episode.
Let $\epcostat{j}$ be the cost of episode $j$; 
clearly $\sum_{j=1}^k \epcostat{j} \leq \budget$. 
We define the multi-fidelity cumulative regret as follows.
\begin{definition}[Cumulative regret]\label{def:cumregret}
  The cumulative regret of policy $\policy$ under budget $\budget$ is
  \begin{align}\label{eq:cumregret}
    \cumreg(\policy,\budget) = \frac{\budget}{\costof{\targetfid}} \tarf^* - \sum_{j=1}^k \reward(\epselectedat{j})
  \end{align}
\end{definition}
Intuitively\footnote{Note that our notion of cumulative regret is different from the multi-fidelity regret (Eq.~(2)) of \cite{kandasamy2016gaussian}. Although both definitions reduce to the classical single-fidelity regret \citep{srinivas10gaussian} when $m=1$, \defref{def:cumregret} has a simpler form and intuitive physical interpretation.}, \defref{def:cumregret} characterizes the difference in the cumulative reward of $\policy$ and the best possible reward gathered under budget $\budget$.

\subsection{Regret Analysis}\label{sec:analysis:regret}
In the following, we establish a bound on the cumulative regret of \algname, as a function of the mutual information between the target fidelity function and the actions attained by the algorithm.
\begin{theorem}\label{thm:general:datadependent-bound}
  Assume that \algname terminates in $k$ episodes, and w.h.p., the cumulative regret incurred by \sfgpopt is upper bounded by
  $\constgeneral_1 \sqrt{\budget {\gamma_m}}$, where $C_1$ is some constant independent of $\budget$, and 
  $\gamma_m$ denotes the mutual information gathered by the target fidelity actions chosen by \sfgpopt (equivalently by \algname). Then,
  w.h.p, the cumulative regret of \algname (\algref{alg:main}) satisfies
  \begin{align*}
    \cumreg(\algname,\budget) \leq \constgeneral_1 \sqrt{\budget {\gamma_m}} + \constgeneral_2 \alpha_\budget \gamma_\eplow
  \end{align*}
  where $\alpha_\budget = \max_{B \leq \budget} \alpha(B)$,  $C_2$ is some constant independent of $\budget$,  
  and $$\gamma_\eplow = \sum_{j=1}^k\condinfgain{\bobs_{\eplow}^{(j)} }{\bobs_{\epselected}^{(1:j-1)}}$$ denotes the mutual information gathered by the low fidelity actions when running \algname.
\end{theorem}
The proof of \thmref{thm:general:datadependent-bound} is provided in the Appendix. Similarly with the single-fidelity case, a desirable asymptotic property of a multi-fidelity optimization algorithm is to be no-regret, i.e., $\lim_{\budget\rightarrow \infty} \cumreg(\policy,\budget)/\budget \rightarrow 0$.
If we set $\alpha(B) = \littleO{\sqrt{B}}$, then \thmref{thm:general:datadependent-bound} reduces to
$\cumreg(\algname,\budget) \leq \constgeneral_1 \sqrt{\budget {\gamma_m}} + \constgeneral_2 \gamma_{\eplow} \cdot \littleO{\sqrt{\budget}}$.
Clearly, \algname is no-regret as $\budget\rightarrow \infty$.

Furthermore, let us compare the above result with the regret bound of the single-fidelity GP optimization algorithm \sfgpopt. By the assumption of \thmref{thm:general:datadependent-bound}, we know $\cumreg(\sfgpopt,\budget) = \bigO{\sqrt{\budget \gamma_m'}}$, where $\gamma_m'$ is the information gain of all the (target fidelity) actions by running \sfgpopt for $\floor{\budget/\costof{\targetfid}}$ rounds.
When the low fidelity actions are very informative about the $\tarf$ and have much lower costs than $\costof{\targetfid}$ (hence larger $\gamma_\eplow$), the less likely \algname will focus on exploring the target fidelity, i.e., $\gamma_m \leq \gamma_m'$, and hence \algname becomes more advantageous to \sfgpopt. The implication of \thmref{thm:general:datadependent-bound} is similar in spirit to the regret bound provided in \cite{kandasamy2016gaussian}, however, our results apply to a much broader class of optimization strategies, as suggested by the following corollary.

\begin{corollary}\label{cor:gpucb}
  Let $\alpha(B)=\littleO{\sqrt{B}}$. Running \algname with subroutine \sfgpucb, \sfest, or \sfgpmi in the optimization phase is no-regret.
\end{corollary}


\section{Experiments}\label{sec:exp}
In this section, we empirically evaluate our algorithm on 3 synthetic test function optimization tasks and 2 practical optimization problems.
\subsection{Experimental Setup} \label{sec:exp:setup}
To model the relationship between a low fidelity function $\utility_{i}$ and the target fidelity function $\tarf$, we use an additive model. Specifically, we assume that $\utility_i = \tarf + \noise_i$ for all fidelity levels $i < m$ where $\noise_i$ is an unknown function characterizing the error incurred by a lower fidelity function. We use Gaussian processes to model $\tarf$ and $\noise_i$. Since $\tarf$ is embedded in every fidelity level, we can use an observation from any fidelity to update the posterior for \emph{every} fidelity level. We use square exponential kernels for all the GP covariances, with hyperparameter tuning scheduled periodically during optimization. We keep the same experimental setup for \mfgpucb as in \citet{kandasamy2016gaussian}.

For all our experiments, we use a total budget of 100 times the cost of target fidelity function call $\tarf$. When the optimal value $\tarf^*$ for $\tarf$ is known, we compare simple regrets. Otherwise, we compare simple rewards. 

\subsection{Compared Methods}
Our framework is general and we could plug in different single fidelity Bayesian optimization algorithms for the \sfgpopt procedure in Algorithm \ref{alg:main}. In our experiment, we choose to use \sfgpucb as one instantiation. We compare with \mfgpucb \citep{kandasamy2016gaussian} and \sfgpucb \citep{srinivas10gaussian}.

\subsection{Synthetic Examples}
\begin{figure*}[t]
  \centering
  \begin{subfigure}[b]{.32\textwidth}
    \centering
    {
      \includegraphics[trim={0pt 0pt 0pt 0pt}, width=\textwidth]{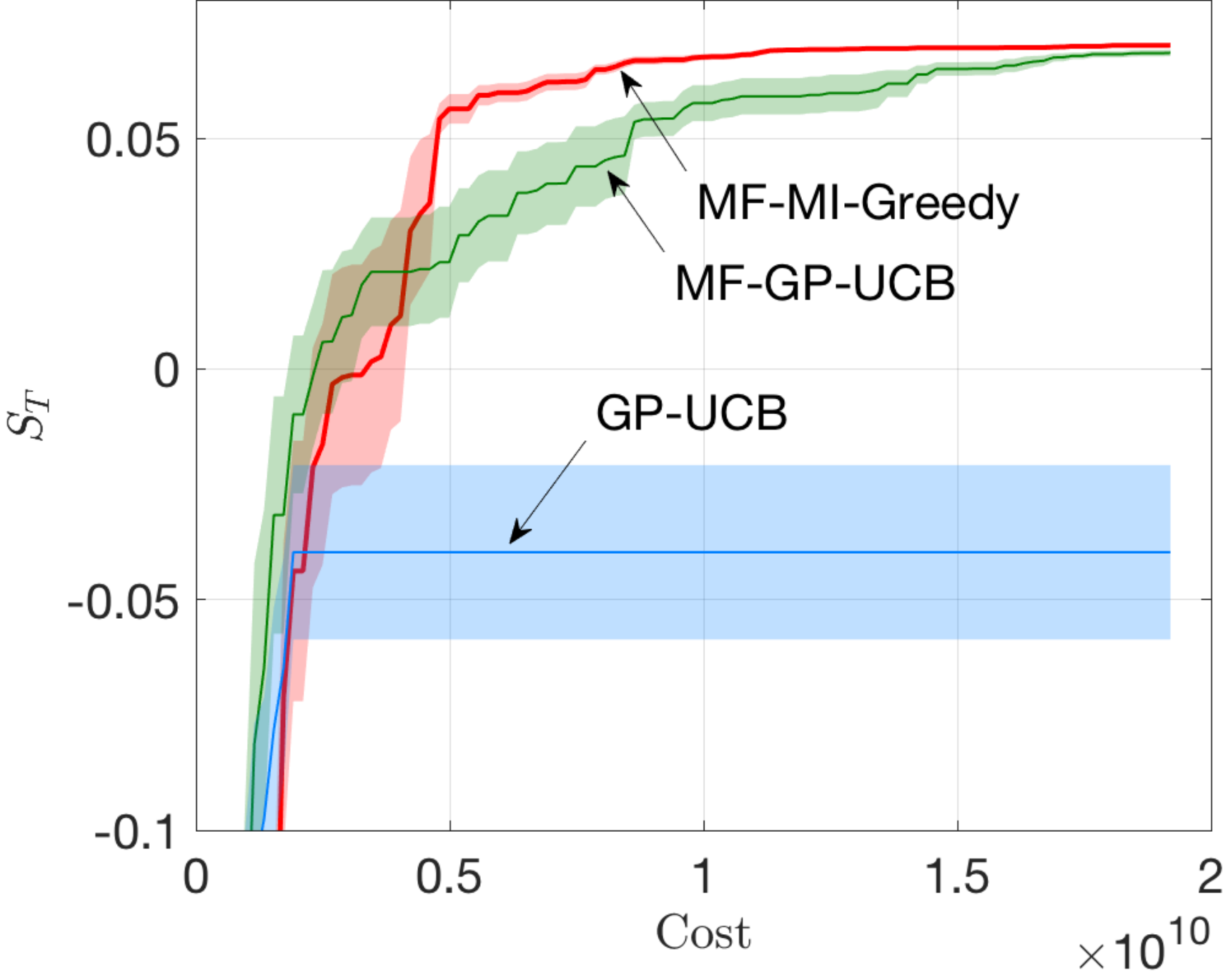}
      \caption{M.L. with grid cost.}
      \label{fig:real:supernova:grid}
    }
  \end{subfigure}
  \begin{subfigure}[b]{.32\textwidth}
    \centering
    {
      \includegraphics[trim={0pt 0pt 0pt 0pt}, width=\textwidth]{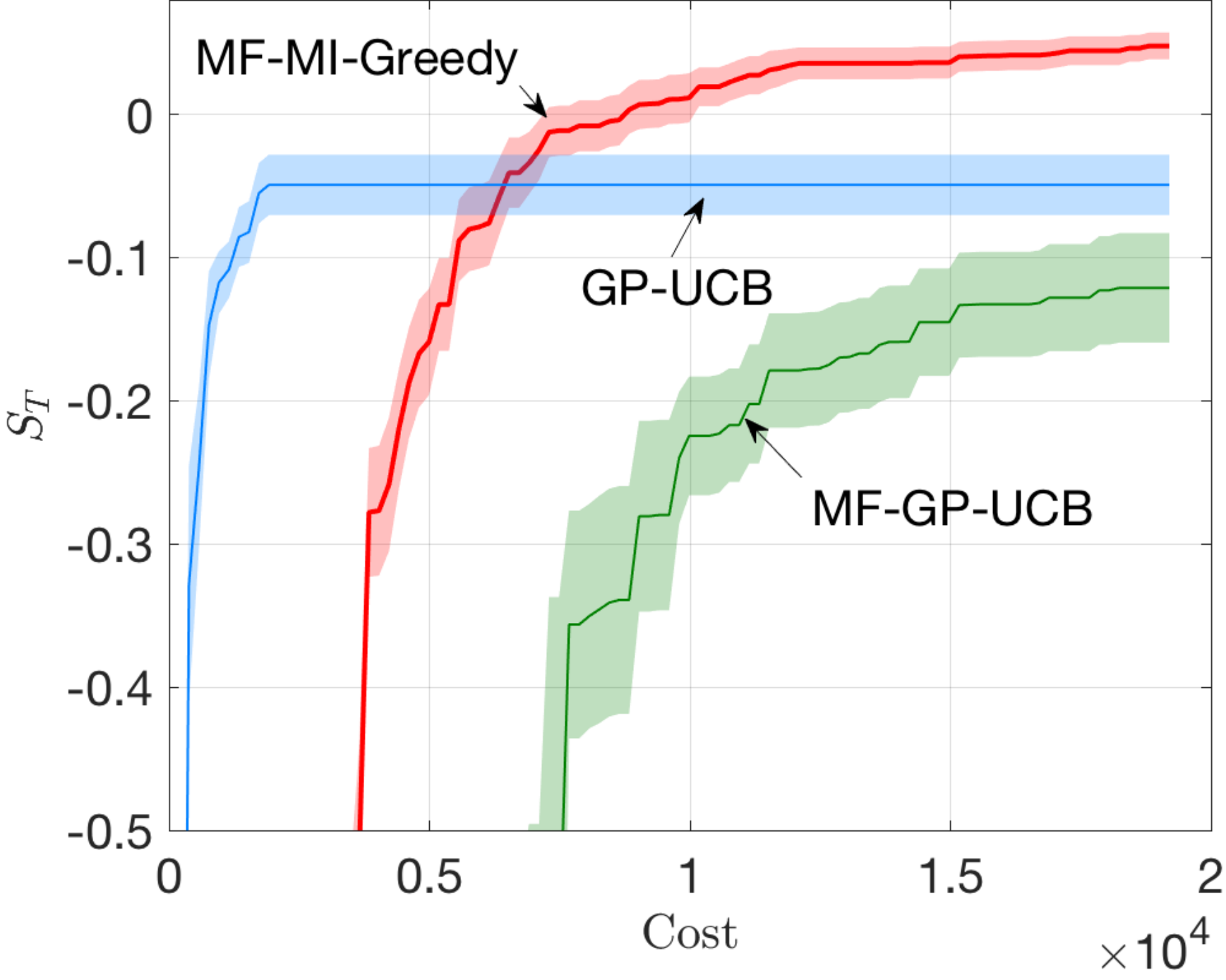}
      \caption{M.L. with observation cost.}
      \label{fig:real:supernova:nogrid}
    }
  \end{subfigure}
  \begin{subfigure}[b]{.32\textwidth}
    \centering
    {
      \includegraphics[trim={0pt 0pt 0pt 0pt}, width=\textwidth]{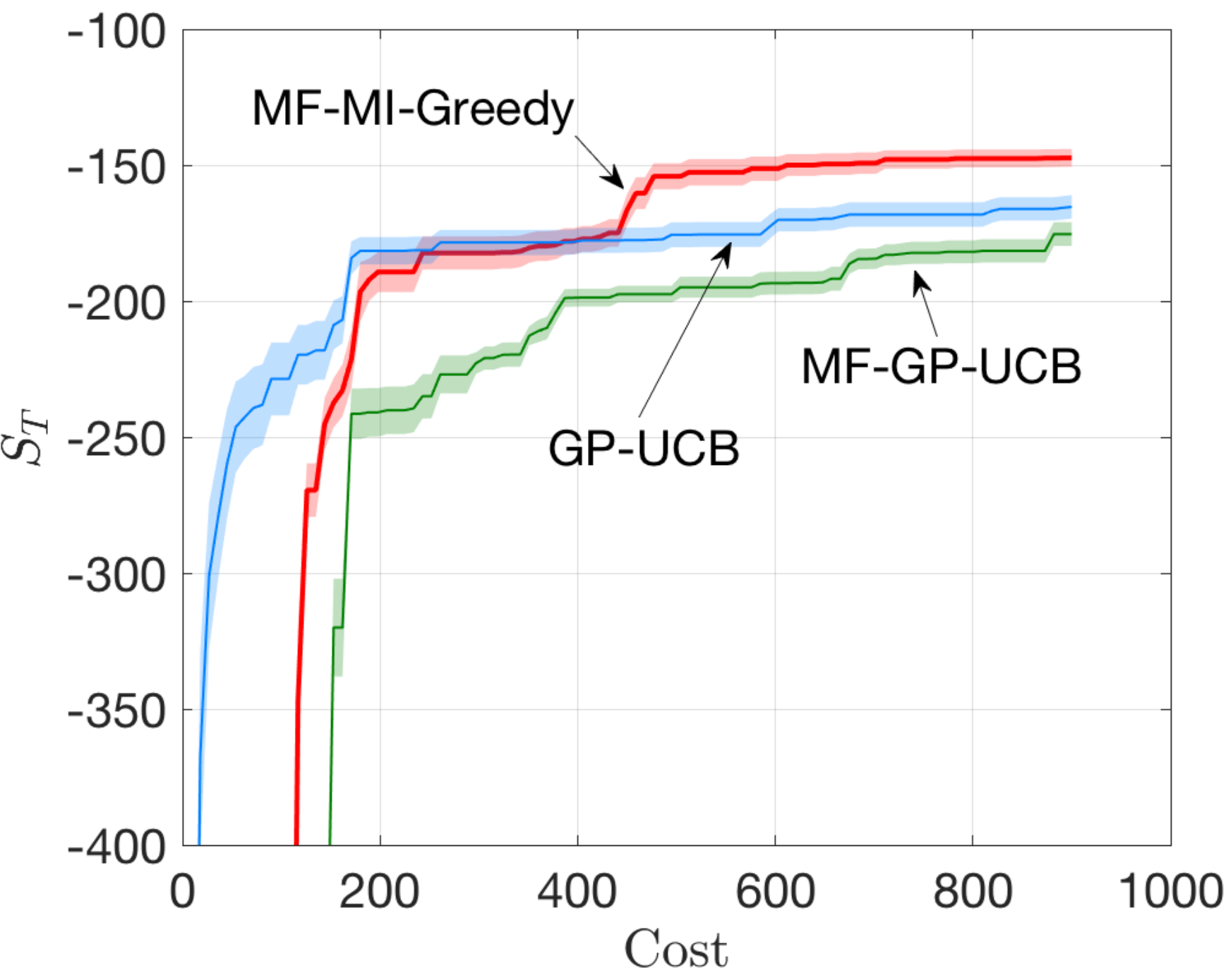}
      \caption{Nanophotonic FOM.}
      \label{fig:real:fom}
    }
  \end{subfigure}
    \caption{Two cost settings for maximum likelihood inference task and the task of optimizing FOM for nanophotonic structures. Error bar shows one standard error over 20 runs for each experiment.}
    \label{fig:exp:real}
\end{figure*}

We first evaluate our algorithm on three synthetic datasets, namely (a) Hartmann 6D, (b) Currin exponential 2D and (c) BoreHole 8D \citep{kandasamy2016gaussian}.
We follow the setup used in \citet{kandasamy2016gaussian} to define lower fidelity functions, while we use a different definition of lower fidelity costs. We emphasize that in synthetic settings, the artificially defined costs do not have practical meanings, as function evaluation costs do not differ across different fidelity levels. Nevertheless, we set the cost of the function evaluations (monotonically) according to the fidelity levels, and present the results in \figref{fig:exp:real}
The $x$-axis represents the expended budget and the $y$-axis represents the smallest simple regret. The error bars represent one standard error over 20 runs of each experiment.

Our method \algname is generally competitive with \mfgpucb. A common issue is its simple regrets tend to be larger at the beginning. A cause for this behavior may be the parameters controlling the termination conditions early on are not tuned optimally, which leads to over exploration in regions that do not reveal much information on where the function optimum lies.

\begin{figure*}[t]
  \centering
  \begin{subfigure}[b]{.32\textwidth}
    \centering
    {
      \includegraphics[trim={0pt 0pt 0pt 0pt}, width=\textwidth]{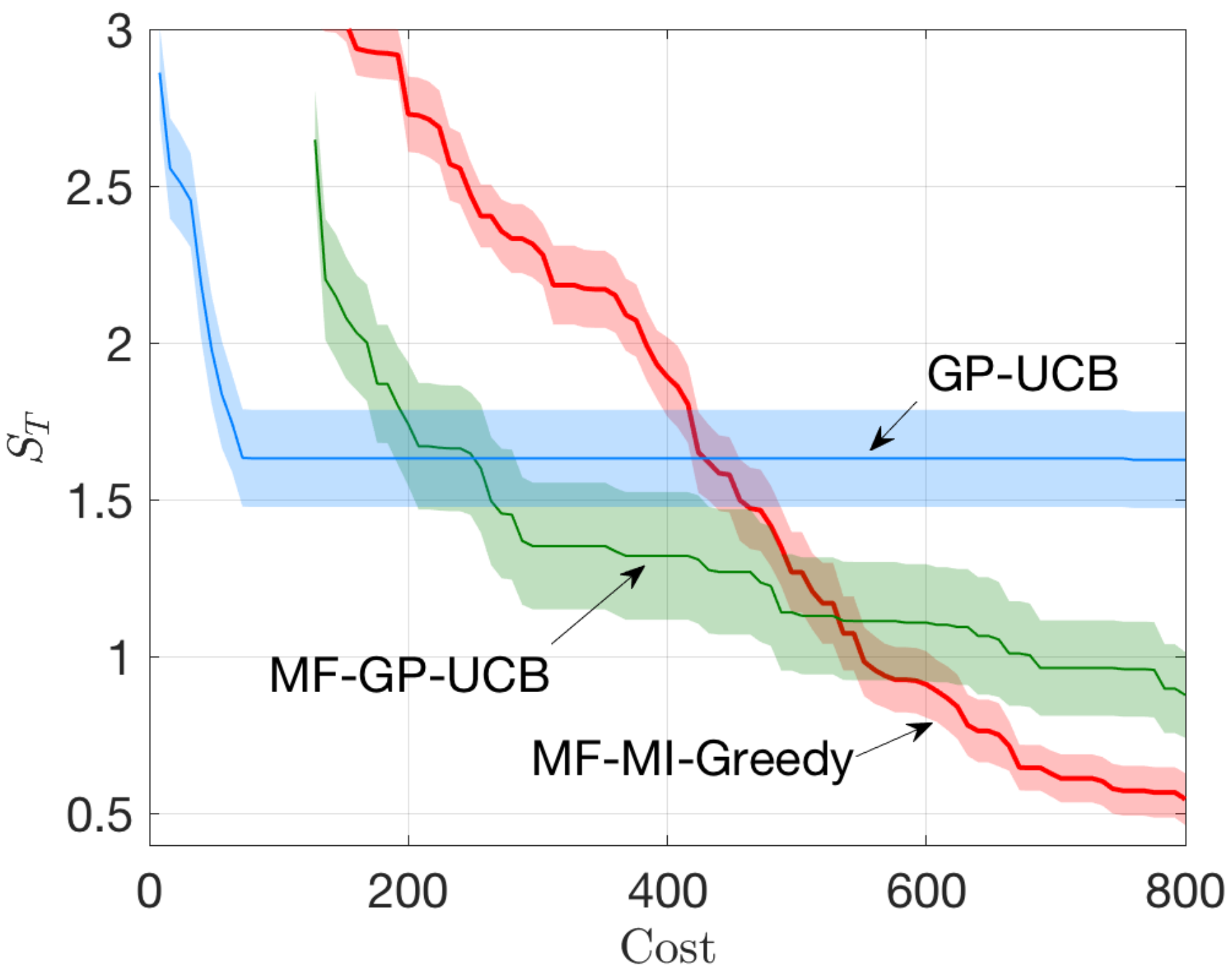}
      \caption{Hartmann 6D]}
      \label{fig:exp:sync:dataset1}
    }
  \end{subfigure}
  \begin{subfigure}[b]{.32\textwidth}
    \centering
    {
      \includegraphics[trim={0pt 0pt 0pt 0pt}, width=\textwidth]{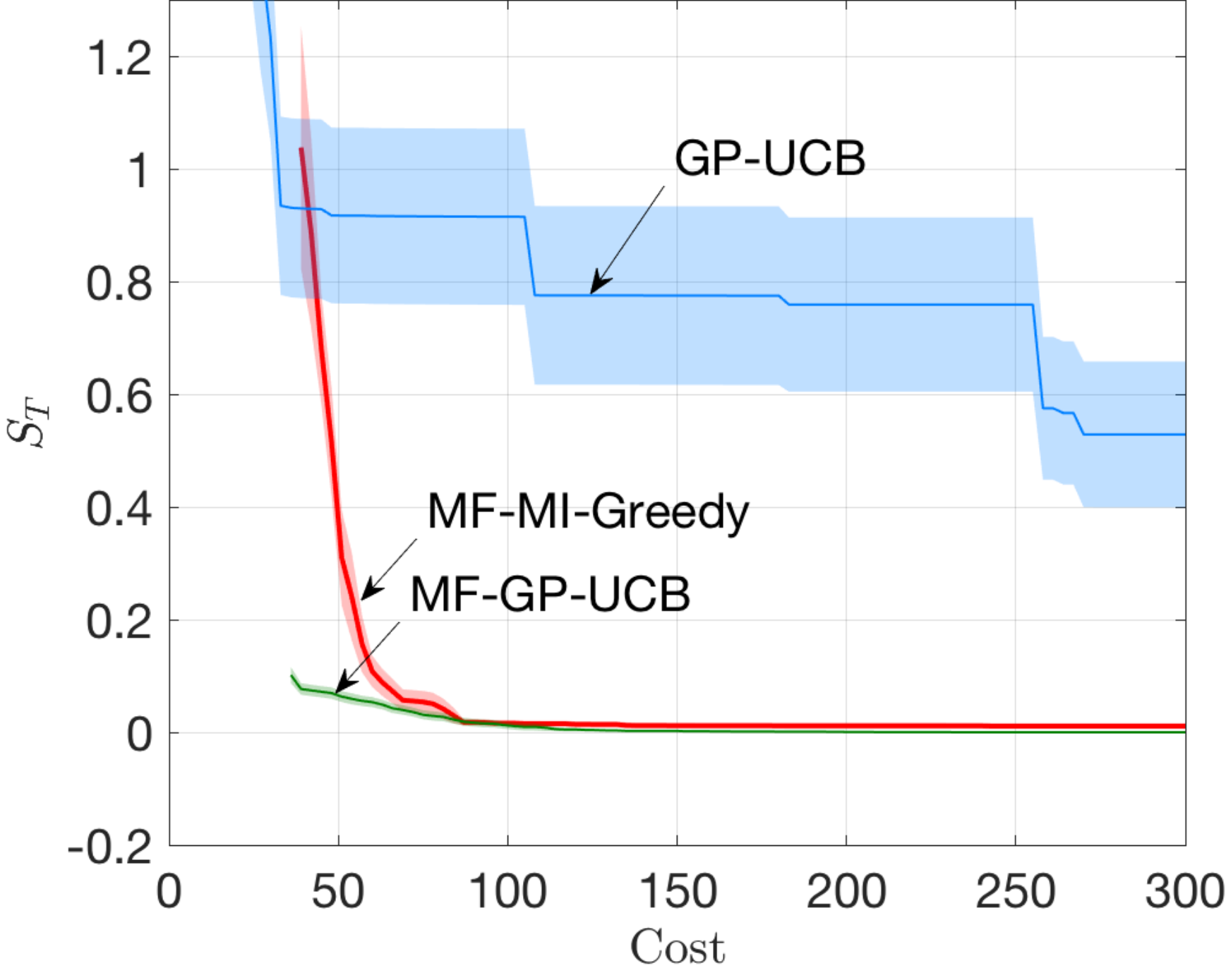}
      \caption{Currin Exp 2D}
      \label{fig:exp:sync:dataset2}
    }
  \end{subfigure}
  \begin{subfigure}[b]{.32\textwidth}
    \centering
    {
      \includegraphics[trim={0pt 0pt 0pt 0pt}, width=\textwidth]{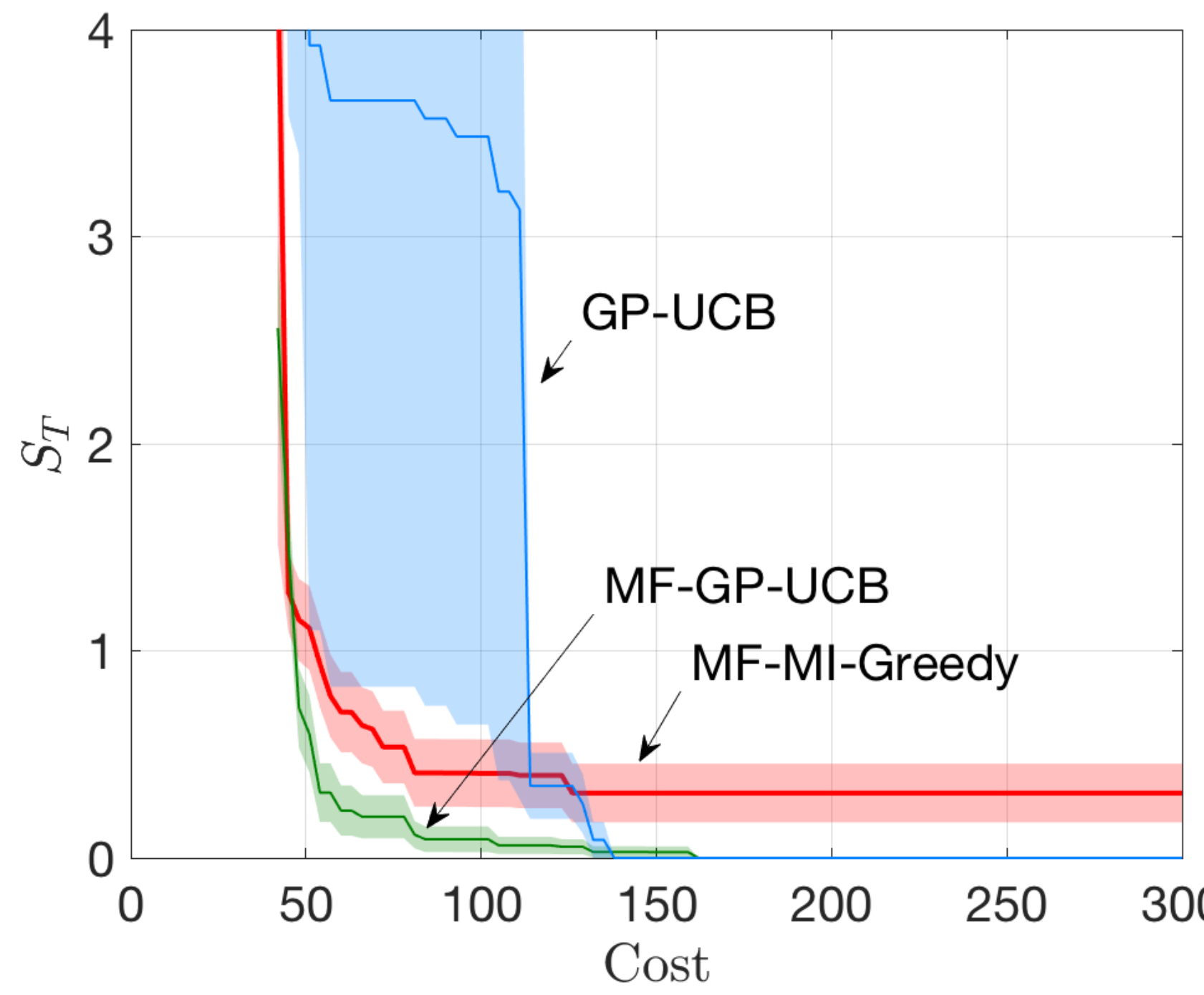}
      \caption{BoreHole 8D}
      \label{fig:exp:sync:dataset3}
    }
  \end{subfigure}

  \caption{Synthetic datasets. For the Hartmann 6D dataset, costs = [1, 2 ,4, 8]; for Currin Exp 2D, costs = [1, 3]; for BoreHole 8D, costs = [1, 2]. Error bar shows one standard error over 20 runs for each experiment.}
    \label{fig:exp:results}
\end{figure*}
\subsection{Real Experiments}
We test our methods on two real datasets: maximum likelihood inference for cosmological parameters and experimental design for material science.

\subsubsection{Maximum Likelihood Inference for Cosmological Parameters}
The first real experiment is to perform maximum likelihood inference on 3 cosmological parameters, the Hubble constant $H_0\in (60, 80)$, the dark matter fraction $\Omega_M\in (0, 1)$ and the dark energy fraction $\Omega_A\in (0, 1)$. It thus has a dimensionality of 3. The likelihood is given by the Roberson-Walker metric, which requires a one-dimensional numerical integration for each point in the dataset from \cite{davis2007scrutinizing}. In \cite{kandasamy2017multi}, the authors set up two lower fidelity functions by considering two aspects of computing the likelihood: (i) how many data points (denoted by $N$) are used, and (ii) what is the discrete grid size (denoted by $G$) for performing the numerical integration. The range for these two parameters are $N\in [50, 192]$ and $G\in [10^2, 10^6]$.
We follow the fidelity levels selected in \cite{kandasamy2017multi} which correspond to two lower fidelities with $(N_1, G_1) = (97, 2.15\times 10^3)$, $(N_2, G_2) = (145, 4.64\times 10^4)$ and the target fidelity with $(N_3, G_3) = (192, 10^6)$. Costs are defined as the product of $N$ and $G$. 

Upon further investigation, we find that the grid sizes selected above for performing numerical integration do not affect the final integral values, i.e. the grid size for the lowest fidelity $G_1 = 2.15 \times 10^3$ is fine enough to compute an approximation to the integration as using the grid size for the target fidelity. So costs taking into consideration the integration grid sizes are not an accurate characterization of the true computation costs. As a result, we propose a different cost definition that depends only on how many data points are used to compute the likelihood, i.e. the new costs for the 3 functions are $(97, 145, 192)$, respectively. 

The results using the original cost definition are shown in Figure \ref{fig:real:supernova:grid}. Note for this task we do not know the optimal likelihood, so we report the best objective value so far (simple rewards) in the $y$-axis. Our method \algname (red) outperforms both baselines. The results using the new cost definition are shown in Figure \ref{fig:real:supernova:nogrid}. Our method obtains a consistent high likelihood when the cost structure changes. However, \mfgpucb's quality degrades significantly, which implies that it is sensitive to how the costs among fidelity levels are defined. These two set of results demonstrate the robustness of our method against costs, which is a desirable property as inaccuracy in cost estimates is inevitable in practical applications. 

\subsubsection{Experimental Design for Optimizing Nanophotonic Structures}

The second experiment is motivated by a material science task of designing nanophotonic structures with desired color filtering property \citep{fleischman2017hyper}.
A nanophotonic structure is characterized by the following 5 parameters:  
mirror height, film thickness, mirror spacing, slit width, and oxide thickness. For each parameter setting, we use a score, commonly called a figure-of-merit (FOM), to represent how well the resulting structure satisfies the desired color filtering property. By minimizing FOM, we hope to find a set of high-quality design parameters.

Traditionally, FOMs can only be computed through the actual fabrication of a structure and tests its various physical properties, which is a time-consuming process. Alternatively, simulations can be utilized to estimate what physical properties a design will have. By solving a variant of the Maxwell's equations, we could simulate the transimission of light spectrum and compute FOM from the spectrum. We collect three fidelity level data on 5000 nanophotonic structures. What distinguishes each fidelity is the mesh size we use to solve the Maxwell's equations. Finer meshes lead to more accurate results. Specifically, lowest fidelity uses a mesh size of $3\text{nm}\times 3\text{nm}$, the middle fidelity $2\text{nm}\times 2\text{nm}$ and the target fidelity $1\text{nm}\times 1\text{nm}$. The costs, simulation time, are inverse proportional to the mesh size, so we use the following costs [1, 4, 9] for our three fidelity functions.

Figure \ref{fig:real:fom} shows the results of this experiment. As usual, the $x$-axis is the cost and $y$-axis is negative FOM. After a small portion of the budget is used in initial exploration, \algname (red) is able to arrive at a better final design compared with \mfgpucb and \sfgpucb.



\section{Conclusion}
In this paper, we investigated the multi-fidelity Bayesian optimization problem, and proposed a general, principled framework for addressing the problem. We introduced a simple, intuitive notion of regret, and showed that our framework is able to lift many popular, off-the-shelf single-fidelity GP optimization algorithms to the multi-fidelity setting, while still preserving their original regret bounds. We demonstrated the performance of our proposed algorithm on several synthetic and real datasets.

\section{Acknowledgments}
This work was supported in part by NSF Award \#1645832, Northrop Grumman, Bloomberg, and a Swiss NSF Early Mobility Postdoctoral Fellowship.


\bibliographystyle{icml2018}
\bibliography{reference}


\clearpage
\appendix
\section{Proofs for \secref{sec:analysis}}

\subsection{Proofs of \thmref{thm:general:datadependent-bound}}

\begin{proof}[Proof of \thmref{thm:general:datadependent-bound}]
  Assume that \algname terminates within $k$ episodes. Let us use $\epselectedat{1}, \dots, \epselectedat{k}$ to denote the sequence of actions selected by \algname, where $\epselectedat{j} := \eplowat{j} \cup \{\tarselectedat{j}\}$ denotes the sequence of actions selected at the $j^{\text{th}}$ episode.
  Further, let $\epcostat{j}$ be the cost of the $j^{\text{th}}$ episode, and $\eplowcostat{j} = \epcostat{j} - \costof{\targetfid}$ the cost of lower fidelity actions of the $j^{\text{th}}$ episode. The budget allocated for the target fidelity is $k\lambda_m = \budget - \sum_{j=1}^k\eplowcostat{j}$.
  By definition of the cumulative regret (Eq.~\eqref{eq:cumregret}), we get
  \begin{align}
    \cumreg(\policy,\budget)
    &= {\frac{\budget}{\costof{\targetfid}}} \tarf^* - \sum_{j=1}^k \reward(\eplowat{j} \cup \{\tarselectedat{j}\}) \nonumber \\
    &= \frac{\budget}{\costof{\targetfid}} \tarf^* - \paren{\sum_{j=1}^k \cancelto{0}{\reward(\eplowat{j})} + \sum_{j=1}^k \reward(\tarselectedat{j})} \nonumber \\
    &= \frac{\budget}{\costof{\targetfid}} \tarf^* -  \sum_{j=1}^{k} \tarf\paren{\ex^{\paren{j}}} \nonumber \\
    &= \paren{\frac{\budget}{\costof{\targetfid}}-k} \tarf^* +  \sum_{j=1}^{k} \paren{\tarf^* - \tarf\paren{\ex^{\paren{j}}}} \nonumber \\
    &= \frac{\tarf^*}{\costof{\targetfid}} \cdot \sum_{j=1}^k{\eplowcostat{j}}  +  \sum_{j=1}^{k} \paren{\tarf^* - \tarf\paren{\ex^{\paren{j}}}} \label{eq:general-regret-intermediate}
  \end{align}
  The first term on the R.H.S. of Eq.~\eqref{eq:general-regret-intermediate} represents the regret incurred from exploring the lower fidelity actions, while the second term represents the regret from the target fidelity actions (chosen by \sfgpopt).

  According to the stopping condition of \algref{alg:explorelf} at Line \ref{alg:explorelf:ln:infgaincheck},
  we know that when \explorelf terminates at episode $j$, the selected low fidelity actions $\eplow$ satisfy
  \begin{align*}
    \frac{\condinfgain{\bobs_{\eplow}^{(j)} }{\bobs_{\epselected}^{(1:j-1)}}}{\eplowcostat{j}} \geq \beta_j,
  \end{align*}
  where $\bobs_{\epselected}^{(1:j)} := \bigcup_{u=1}^{j} \bobs_{\epselected}^{(u)}$ denotes the observations obtained up to episode $j$, and $\beta_j$ specifies the stopping condition of \explorelf at episode $j$.
  Therefore
  \begin{align}
    \sum_{j=1}^k{\eplowcostat{j}}
    &\leq \sum_{j=1}^k\frac{\condinfgain{\bobs_{\eplow}^{(j)} }{\bobs_{\epselected}^{(1:j-1)}}}{\beta_j} \label{eq:app:sumeplowcostbound}\\
    &\stackrel{(a)}{\leq} \alpha_{\budget} \sum_{j=1}^k\condinfgain{\bobs_{\eplow}^{(j)} }{\bobs_{\epselected}^{(1:j-1)}} \nonumber \\
    &= \alpha_{\budget}\gamma_\eplow \nonumber
  \end{align}
  where step (a) is because $\alpha_\budget = \max_{B} \alpha(B) > \frac{1}{\beta_j}$ for $j\in[k]$. Recall that $\beta_1 = \frac{1}{\littleO{\sqrt{\budget}}}$. Therefore, 
  \begin{align}
    \frac{\tarf^*}{\costof{\targetfid}} \cdot \sum_{j=1}^k{\eplowcostat{j}} \leq \frac{\tarf^*}{\costof{\targetfid}} \cdot \alpha_\budget \gamma_\eplow 
    . \label{eq:general-cost-lf-ub}
  \end{align}
  Note that the second term of Eq.~\eqref{eq:general-regret-intermediate} is the regret of \algname on the target fidelity. Since all the target fidelity actions are selected by the subroutine \sfgpopt, by assumption, we know $\sum_{j=1}^{k} \paren{\tarf^* - \tarf\paren{\ex^{\paren{j}}}} = \sqrt{C \gamma_m k\costof{\targetfid}} \leq \sqrt{C \gamma_m \budget}$. Combining this with Eq.~\eqref{eq:general-cost-lf-ub} completes the proof.
\end{proof}

\subsection{Proof of \corref{cor:gpucb}}
To show that running \algname with subroutine \sfgpucb \citep{srinivas10gaussian}, \sfest \citep{wang2016optimization}, or \sfgpmi \citep{contal2014gaussian} in the optimization phase is no-regret, it suffices to show that the candidate subroutines \sfgpucb, \sfest, and \sfgpmi satisfy the assumption on \sfgpopt as provided in \thmref{thm:general:datadependent-bound}. From the references above we know that the statement is true.





\end{document}